\documentclass[lettersize,journal]{IEEEtran}
\usepackage{amsmath,amsfonts}
\usepackage{algorithmic}
\usepackage{algorithm}
\usepackage{array}
\usepackage[caption=false,font=normalsize,labelfont=sf,textfont=sf]{subfig}
\usepackage{textcomp}
\usepackage{stfloats}
\usepackage{url}
\usepackage{verbatim}
\usepackage{graphicx}
\usepackage{cite}
\usepackage{float}
\usepackage{booktabs}
\usepackage{multirow}
\usepackage{balance}
\begin{document}

\title{Tuning a SAM-Based Model\\ with Multi-Cognitive Visual Adapter\\ to Remote Sensing Instance Segmentation}

\author{Linghao Zheng,~\IEEEmembership{Student Member,~IEEE}, Xinyang Pu,~\IEEEmembership{Graduate Student Member,~IEEE}, Feng Xu,~\IEEEmembership{Senior Member,~IEEE}
  % <-this % stops a space
  \thanks{The authors are with the Key Laboratory for Information Science of Electromagnetic Waves (Ministry of Education), School of Information Science and Technology, Fudan University, Shanghai 200433, China. (Corresponding author: Feng Xu, fengxu@fudan.edu.cn)}% <-this % stops a space
  }

\markboth{Journal of \LaTeX\ Class Files,~Vol.~14, No.~8, August~2021}%
{Shell \MakeLowercase{\textit{et al.}}: A Sample Article Using IEEEtran.cls for IEEE Journals}

\IEEEpubid{0000--0000/00\$00.00~\copyright~2021 IEEE}

\maketitle

\begin{abstract}
  The Segment Anything Model (SAM), a foundational model designed for promptable segmentation tasks, demonstrates exceptional generalization capabilities, 
  making it highly promising for natural scene image segmentation. However, SAM's lack of pretraining on massive remote sensing images and its interactive structure limit its automatic mask prediction capabilities. 
In this paper, a Multi-Cognitive SAM-Based Instance Segmentation Model (MC-SAM SEG) is introduced to employ SAM on remote sensing domain. 
The SAM-Mona encoder utilizing the Multi-cognitive Visual Adapter (Mona) is conducted to facilitate SAM's transfer learning in remote sensing applications. 
The proposed method named MC-SAM SEG extracts high-quality features by fine-tuning the SAM-Mona encoder along with a feature aggregator. 
Subsequently, a pixel decoder and transformer decoder are designed for prompt-free mask generation and instance classification.
The comprehensive experiments are conducted on the HRSID and WHU datasets for instance segmentation tasks on Synthetic Aperture Radar (SAR) images and optical remote sensing images respectively. The evaluation results indicate the proposed method surpasses other deep learning algorithms and verify its effectiveness and generalization.
\end{abstract}


\section{Introduction}
\IEEEPARstart{R}{emote} sensing images, acquired via aerial or satellite platforms, capture electromagnetic wave and spectral data of Earth's surface features. 
As satellites advance towards large-scale, multi-spectral, high-resolution, and frequent data collection, 
their applications continue to expand \cite{aleissaee2023Transformers}. 
Instance segmentation plays a critical role on fine-grained pixel-level interpretation of remote sensing images\cite{chen2021db}. 
This process involves precise segmentation, instance localization and classification within each pixel, 
thereby providing detailed information that supports Earth observation and various applications\cite{chen2019temporary,pan2018semantic}. 
This technology holds significant potential applications in multiple fields such as marine monitoring, water extraction, building identification, land monitoring, and more\cite{pan2018semantic,chen2022contrastive}.

Previous methods for instance segmentation in remote sensing predominantly rely on traditional deep neural networks with common backbones.  
Waqas et al.\cite{waqas2019isaid} initially achieved instance segmentation in remote sensing images by fine-tuning Mask R-CNN\cite{he2017mask}. 
Su et al.\cite{su2020hq} introduced Cascade Mask R-CNN\cite{cai2018cascade} for remote sensing image segmentation, 
proposing the High-Quality Instance Segmentation Network (HQ-ISNet) model. 
Zhao et al.\cite{zhao2021synergistic} incorporated various attention mechanisms, 
developing a collaborative attention-based method for SAR remote sensing image instance segmentation.
Gao et al.\cite{gao2021anchor} introduced an anchor-free model, enhancing performance with the addition of centroid distance loss. 
Fan et al.\cite{fan2022efficient} utilized the Swin Transformer for ship feature extraction.
Traditional deep learning methods have extensively designed features for remote sensing image characteristics. 
However, due to the insufficient feature extraction capabilities of backbone networks, these methods generally lack robustness and face accuracy bottlenecks.

Recently, Segment Anything Model (SAM) \cite{kirillov2023segment} has pushed the boundaries of segmentation, 
demonstrating significant generalization and zero-shot inference capabilities.  
The decoupled architecture of SAM grants it computational efficiency, 
which is crucial for large-scale remote sensing applications that involve processing vast amounts of data\cite{osco2023segment}.
Consequently, SAM has been applied on various computer vision downstream tasks in remote sensing field.
Chen et al. \cite{chen2024rsprompter} proposed RSPrompter, a prompt-based learning approach utilizing SAM for instance segmentation in remote sensing imagery. 
Yan et al. \cite{yan2023ringmo}  introduced RingMo-SAM for multi-modal remote sensing image segmentation, 
designing a prompt encoder to embed features from multi-modal remote sensing data to enhance SAM's segmentation accuracy. 
These methods significantly improve segmentation accuracy, but several limitations still exist and constrain further performance enhancements of SAM in remote sensing instance segmentation:
\begin{enumerate}
  \item{The pre-training dataset SA-1B for SAM lacks diversity in remote sensing imagery and some existing methods employ the original SAM image encoder without adapting it for remote sensing through fine-tuning, which leads to the omission of important features such as unique scattering characteristics, dense instance layouts, and varied scales, especially in SAR images. 
  Consequently, SAM does not exhibit satisfactory performance in the task of instance segmentation of remote sensing images.\IEEEpubidadjcol
  }
  \item{As an interactive segmentation method with strong generalization capabilities, 
  SAM relies on high-quality prompt inputs to discard irrelevant features. 
  This reliance makes SAM unsuitable for automatic image interpretation. 
  Consequently, SAM is commonly utilized for the interactive segmentation of remote sensing images, 
  requiring prompt inputs provided by users. } 
  \item{Without input prompts, SAM generates segmentation masks for the most recognizable instances and pixels across the entire image. However, the predicted masks lack category information for the instances, thereby missing classification ability required for specific targets in instance segmentation tasks.}
\end{enumerate} 
This paper introduces the Multi-cognitive Visual Adapter (Mona)\cite{yin2023adapter}into the image encoder of SAM with parameter efficient fine-tuning (PEFT) strategy, and named the image encoder as SAM-Mona. 
The multi-scale cognition capability of Mona facilitates the learning of complex scene segmentation, 
SAR scattering characteristics, and other remote sensing features, thereby achieving higher-quality transfer of SAM into the remote sensing domain. 
The original interactive structure of SAM extracts richer features for generalization and requires prompts to filter out irrelevant information. 
In contrast, the proposed SAM-Mona encoder can automatically extract useful features of remote sensing images.
Building on this, the proposed instance segmentation algorithm, MC-SAM SEG, is able to generate masks and predict categories for each instance without prompt inputs. 
The SAM-Mona encoder and an auxiliary feature aggregator are fine-tuned using PEFT technology to extract features from remote sensing images efficiently, 
while SAM's prompt decoder is discarded. The extracted features are sent to a pixel decoder utilizing Multi-scale Deformable Attention Transformer 
to generate pixel-level results on multiple resolutions, which are then progressively fed into the transformer decoder to generate final masks and predict classes.

In summary, the main contributions are as follows:
\begin{enumerate}
  \item{An instance segmentation algorithm based on SAM for remote sensing images named as MC-SAM SEG is proposed. High-quality transfer of SAM into the remote sensing field for instance segmentation task is achieved through integrating Multi-cognitive Visual Adapter (Mona) into the image encoder of SAM with parameter efficient fine-tuning (PEFT) technology.}
  \item{SAM-Mona encoder is introduced to combine with feature aggregator to extract image features on remote sensing scenes. 
  Subsequently, a pixel decoder and transformer decoder are employed for prompt-free mask generation and category prediction.}
  \item{Extensive experiments on both optical and SAR images demonstrate that our method outperforms other state-of-the-art methods. 
   The proposed MC-SAM SEG achieves an $\mathrm{AP}_{\text {mask }}$of 71.2$\%$ on the building segmentation dataset of optical images named WHU, and an $\mathrm{AP}_{\text {mask }}$of 66.4$\%$ on the SAR ship dataset HRSID, showcasing its promising performance and generalization ability.}
\end{enumerate}

The rest of this paper is arranged as:
Section  \uppercase\expandafter{\romannumeral2} presents an extensive review of prior studies concerning remote sensing image instance segmentation.
Section  \uppercase\expandafter{\romannumeral3} elaborates on the detailed architecture of the proposed method MC-SAM.
Section  \uppercase\expandafter{\romannumeral4} presents a thorough analysis of quantitative and qualitative evaluation results, 
comparing multiple experiments and conducting an ablation study. 
Finally, Section  \uppercase\expandafter{\romannumeral5} concludes the paper.

\section{Related Work}
\subsection{Instance Segmentation in Remote Sensing Scenes}
Instance segmentation algorithms perform pixel-wise segmentation of individual targets in remote sensing images, 
preserving target contours, angles, and actual sizes, thereby demonstrating promising applications. 
However, due to limited datasets and the challenges associated with annotating instance targets, 
research on instance segmentation algorithms for remote sensing images is still in its early stages.

Researchers in the field have explored remote sensing image instance segmentation algorithms to varying extents. 
Chen et al. \cite{chen2019hybrid} proposed a hybrid cascade structure to jointly process multiple tasks. 
Su et al.\cite{su2020hq} introduced Cascade Mask R-CNN\cite{cai2018cascade}  for remote sensing image segmentation and proposed the High-Quality Instance Segmentation Network (HQ-ISNet) model.
Wang et al. \cite{wang2020solo}  introduced the notion of instance categories, assigning categories to each pixel within an instance based on the instance's location and size. 
Vu et al. \cite{vu2021scnet} proposed the Sample Consistency Network (SCNet) architecture to ensure that the IoU distribution of samples during training is consistent with that during inference.
Liu et al. \cite{liu2021catnet} proposed a novel Context Aggregation Network (CATNet) to enhance the feature extraction process.
Gao et al.\cite{gao2021anchor}  introduced an anchor-free model that incorporated centroid distance loss to improve performance. 
Zhang et al.\cite{zhang2021semantic} proposed the Semantic Attention Mechanism (SEA) and Scale-Complementary Network to activate target instances and reduce background noise interference.  
Zhao et al.\cite{zhao2021synergistic} developed the Global Attention Module (GAM), Semantic Attention Module (SAM), and Anchor Attention Module (AAM) for SAR ship instance segmentation using collaborative attention mechanisms.
Wei et al. \cite{wei2022lfg} pioneered the Low-Level Feature Guided Network (LFG-Net) by constructing low-level features to discriminate ships and integrating super-resolution (SR) denoising techniques. 
Zhang et al. \cite{zhang2022htc+} proposed the Hybrid Task Cascade Plus (HTC+) to focus on the specific characteristics of ships in SAR images. 
Cheng et al. \cite{cheng2022masked} presented Masked-attention Mask Transformer (Mask2Former), utilizing masked attention to extract localized features by constraining cross-attention within predicted mask regions.
Zhang et al. \cite{zhang2022full}  introduced the Full-Level Context Squeeze-and-Excitation ROI Extractor (FL-CSE-ROIE) to extract context information of regions of interest (ROI).
Liu et al. \cite{liu2024learning}  introduced the Dense Feature Pyramid Network (DenseFPN), Spatial Context Pyramid (SCP), and Hierarchical RoI Extractor (HRoIE) to enhance network instance segmentation performance.

\subsection{Applying SAM in Segmentation}
SAM \cite{kirillov2023segment} was introduced by Meta in 2023 under the Segment Anything project. 
Researchers aimed to develop a foundational model analogous to those demonstrating strong performance in natural language processing and computer vision, 
with the intent to unify the entire image segmentation task. 
The emergence of SAM provides new ideas and approaches for addressing downstream tasks\cite{chen2023sam,wu2023medical,gong20233dsam,li2023enhancing}.
However, the scarcity of available data in the segmentation field and the different design objectives of the model present significant challenges for this task.

SAM exhibits excellent generalization capabilities in common scenarios like natural images but performs poorly in low-contrast scenes and requires strong prior knowledge in complex scenarios. 
To evaluate SAM's generalization ability in more complex scenes, 
Ji et al. \cite{ji2023sam} quantitatively compared SAM with state-of-the-art models in three occluded scenarios, observing that SAM lacks proficiency in occluded scenarios and may require support from domain-specific prior knowledge.  
He et al. \cite{he2024weakly}  proposed the first method for weakly supervised occluded object segmentation using SAM (WS-SAM), addressing the challenge of segmenting objects that blend with the background using sparse annotation data. 
Keyan Chen et al. introduced RSPrompter \cite{chen2024rsprompter}, a prompt learning method for remote sensing image instance segmentation that leverages the SAM foundation model. RSPrompter aims to learn how to generate prompt inputs for SAM, allowing it to automatically obtain semantic instance-level masks. 
Yan et al. \cite{yan2023ringmo} proposed RingMo-SAM, a foundation model for multimodal remote-sensing image segmentation. 
Pu et al. \cite{pu2024classwise} designed the ClassWise-SAM-Adapter (CWSAM) to adapt SAM for landcover classification on space-borne SAR images.
\begin{figure*}[t]
  \centering
  \includegraphics[width=6in]{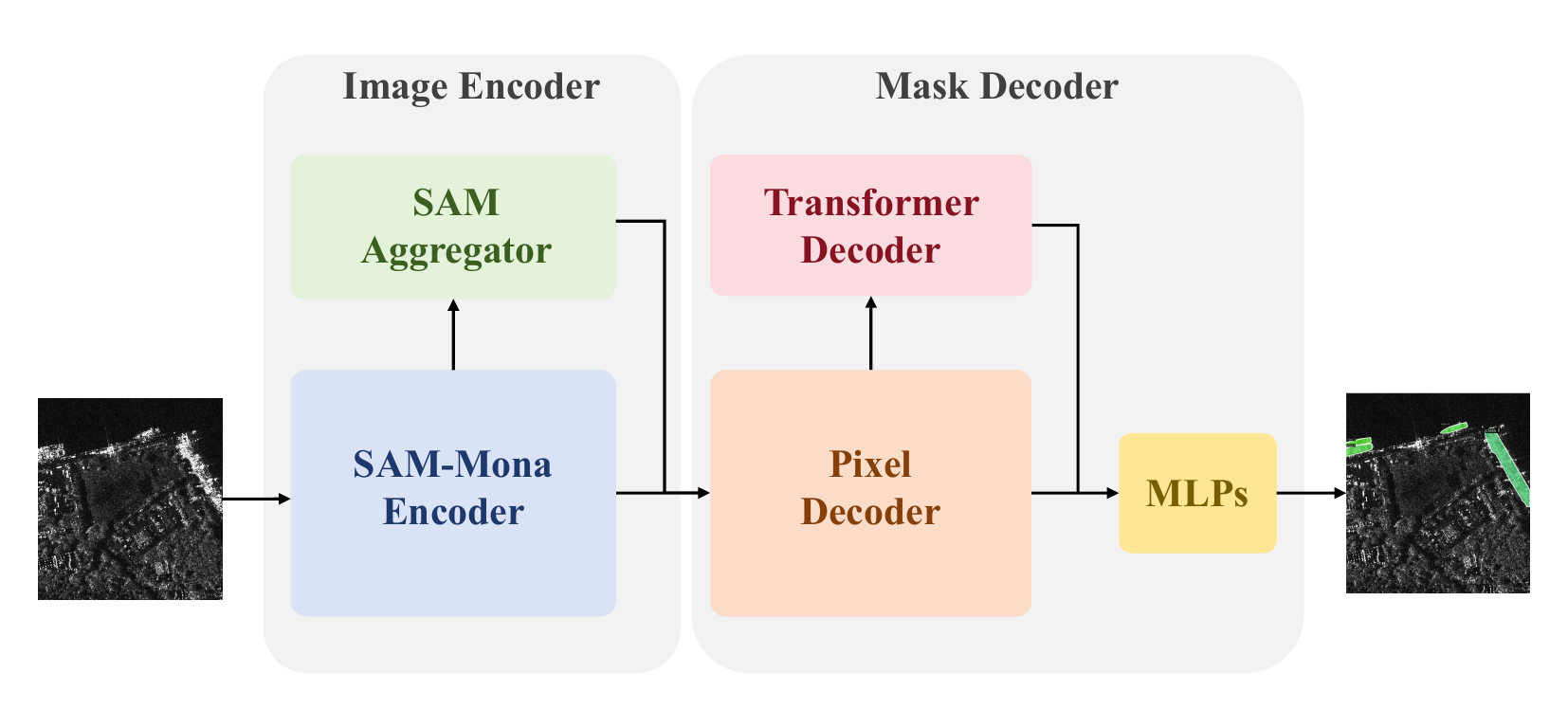}
  \caption{Overview network architecture of MC-SAM SEG, which employs an encoder-decoder architecture.}
  \label{fig_1}
\end{figure*}
\subsection{Parameter-Efficient Fine-Tuning}
Parameter-Efficient Fine-Tuning (PEFT) technology enhances the performance of pre-trained models on new tasks 
by minimizing the number of parameters and computational complexity. This approach allows for rapid adaptation to new tasks using the existing knowledge 
of the model, even with limited computational resources, resulting in efficient transfer learning. 
PEFT improves model effectiveness while reducing both training time and computational expenses\cite{pu2023empirical,xu2023parameter,wang2023towards}.

The current mainstream PEFT methods include Adapter\cite{houlsby2019parameter} , Prefix Tuning\cite{li2021prefix} , and LoRA\cite{hu2021lora} . 
The core idea of the Adapter method is to add residual modules on top of neural network modules and optimize only these residual modules. Since the residual modules contain fewer parameters, the fine-tuning cost is lower \cite{chen2022vision,wang2020k}. 
To challenge the dominant position of full fine-tuning in the computer vision (CV) field, Yin et al.\cite{yin2023adapter} proposed a new fine-tuning paradigm called Mona-tuning. 
Experimental results indicate that Mona-tuning is the only method that surpasses full fine-tuning in semantic segmentation and instance segmentation tasks.
Prefix-Tuning involves constructing task-specific virtual tokens that are inserted before input tokens during training. 
Low-Rank Adaptation (LoRA) approximates parameter updates of a full-rank weight matrix using a learnable low-rank matrix, thereby reducing inference latency and memory consumption.

\section{ METHODOLOGY}
In this section, the proposed method Multi-Cognitive SAM-Based Instance Segmentation Model (MC-SAM SEG) is introduced by three parts. Fig. 1 presents an overview of the proposed framework which employs an encoder-decoder architecture, consisting of an image encoder and a mask decoder, to generate segmentation results automatically without prompts. 
As presented in Section III-A, The image encoder incorporates the SAM-Mona encoder, which integrates Mona-Tuning with SAM, 
enabling it to effectively learn and extract relevant features from remote sensing images. 
In Section III-B, a lightweight feature aggregator is then used to gather multi-scale information from the image encoder and perform lightweight feature fusion using convolutional blocks. 
The aggregated features are subsequently fed into the pixel decoder, which extracts a feature pyramid composed of both low-resolution and high-resolution features. 
These multi-scale features are then fed into different Transformer decoder layers, 
and the decoding results are used for final mask generation and class prediction.
Furthermore, the third section illustrates the detailed transfer learning strategy of the proposed method. 

\subsection{SAM-Mona Encoder}
The image encoder of SAM is a Vision Transformer (ViT) pretrained using Mask Auto Encoder (MAE).
The input image $I \in \mathbb{R}^{h \times w \times 3}$ is cropped into N patches $p_i$, 
which are flattened through a linear layer and concatenated with positional embeddings to obtain N+1 patch embeddings $p e_i$.
These embeddings are then fed into an encoder composed of L standard Transformer blocks to generate intermediate image features $F_{img} \in \mathbb{R}^{h \times w \times c}$. 
The structure of a standard Transformer block is shown in Fig. 2(a).
The patch embeddings, after normalization, pass through multi-head attention to extract comprehensive feature information. 
They are normalized again and then processed by a feedforward network to enhance the model's nonlinear representation capabilities, 
resulting in a feature matrix $F_{img} \in \mathbb{R}^{h \times w \times c}$ of the same dimension as the input. 
Residual connections are added after the multi-head attention and feedforward network to aggregate features, enabling the training of deeper networks.
\begin{figure*}[t]
  \centering
  \includegraphics[width=5.5in]{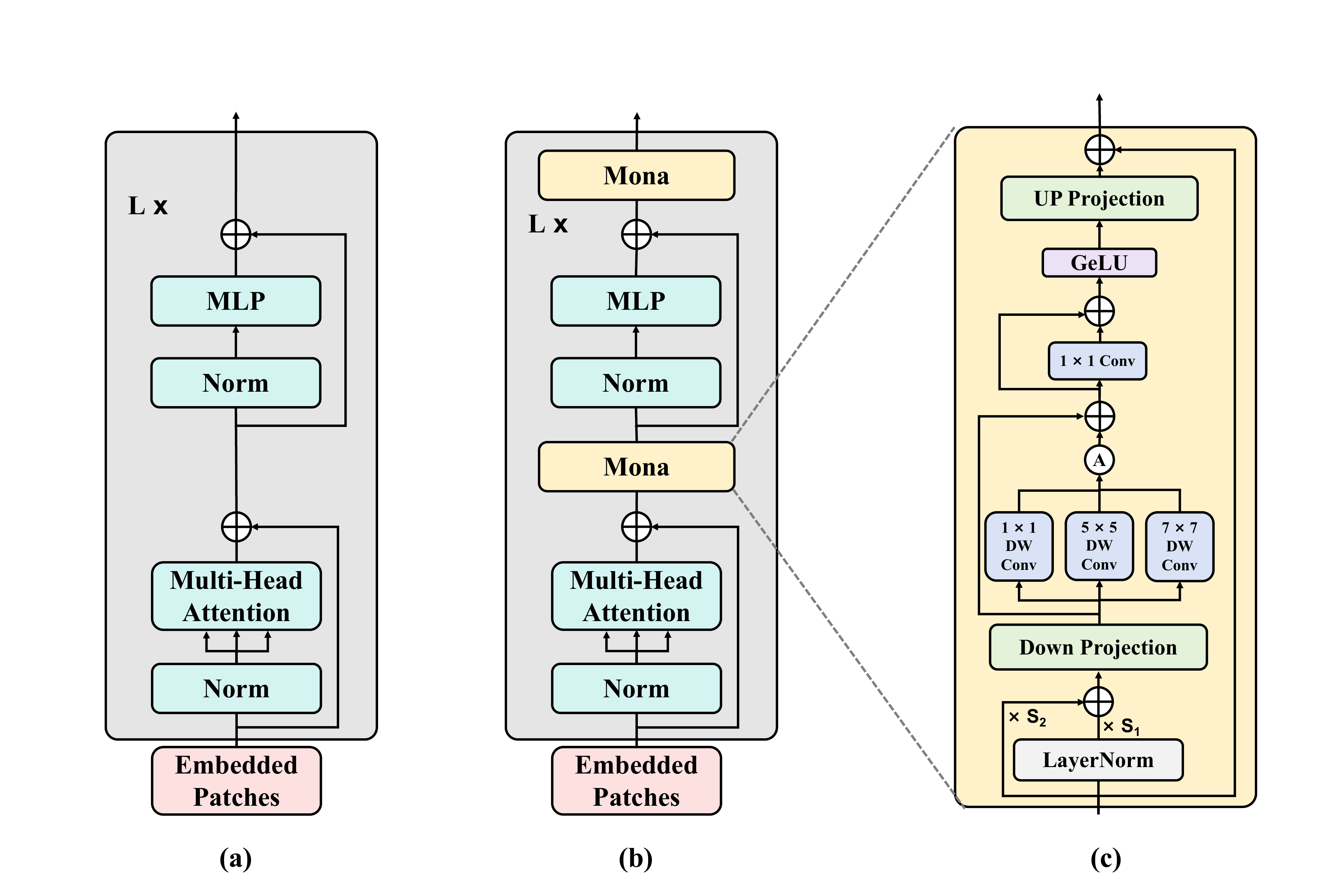}
  \caption{(a) Standard Transformer block. (b) SAM-Mona encoder's Transformer block.(c) Architecture of Mona Adapter. }
  \label{fig_2}
\end{figure*}

The standard transformer block of SAM image encoder inherits its feature extraction capabilities on remote sensing images. 
To enable the learning of specialized semantic information in remote sensing images, 
such as the morphology of complex buildings from an aerial perspective and the diverse scattering characteristics of ground objects in SAR images, 
the Mona adapter is inserted into standard Transformer block of SAM for transfer learning.
Mona \cite{yin2023adapter} is an adapter-based PEFT module with multi-scale cognitive design for visual tasks. 
It can adapt to the larger scale variation range in remote sensing images and focus on local details of instances, 
facilitating a higher quality transfer of SAM to the remote sensing field. 
This approach enables the model to effectively capture and represent the intricate features and variations inherent in remote sensing images.
Mona introduces multiple convolutional filters to enhance cognitive dimensions, 
with the specific structure shown in Fig. 2(c). The computational process in Mona can be described as follows:
\begin{equation}
  \begin{aligned}
  F_i & =\Phi_{\text {LayerNorm}}\left(F_i\right) \times S_1+F_i \times S_2 \\[4pt]
  F_{\text {down}} & =\Phi_{\text {DownProj }}\left(F_i\right) \\[4pt]
  F_{\text {agg}} & =F_{\text {down }}+\text { Average } \sum\left(\Phi_{\text {DWConv } \mathrm{i} \times \mathrm{i}}\left(F_{\text {down}}\right)\right) \\[4pt]
  F_{\text {agg}}^{\prime} & =\mathrm{F}_{\text {agg }}+\Phi_{\text {Conv } 1 \times 1}\left(F_{\text {agg}}\right) \\[4pt]
  F_{\text {up}} & =\Phi_{\text {UpProj }}\left(\Phi_{\text {GeLU }}\left(F_{\text {agg}}^{\prime}\right)\right)
  \end{aligned}
  \end{equation}
Given $F_{i} \in \mathbb{R}^{h \times w \times c}$ as the intermediate features obtained from the Transformer block, 
the features undergo layer normalization $\Phi_{LayerNorm}$ and are scaled by learnable scaling factors \( S_1 \) and \( S_2 \). 
The features are then down-projected to a lower dimension, resulting in  $F_{down} \in \mathbb{R}^{64 \times 64 \times c}$, 
which is computationally more efficient. 
After that, the features are filtered with multiple filters $\Phi_{DWConv}$ of different sizes \( i \) and aggregated to produce $F_{agg} $. 
A \( 1 \times 1 \) convolution is then used for channel aggregation, followed by non-linear transformation $\Phi_{GeLU}$ . 
Finally, the features are up-projected back to the original dimension, resulting in the processed features $F_{up} \in \mathbb{R}^{h \times w \times c}$.

Mona adapter is integrated into SAM's standard Transformer block, 
resulting in the SAM-Mona encoder's Transformer block, and the detailed structure is shown in Fig. 2(b). 
To ensure the ability of processing higher-quality features, 
Two Mona adapter layers are inserted after each of the feature aggregation modules within the transformer block, specifically following both the Multi-Head Attention mechanism and the Multi-Layer Perceptron (MLP).

\subsection{MC-SAM SEG}
\begin{figure*}[t]
  \centering
  \includegraphics[width=6.5in]{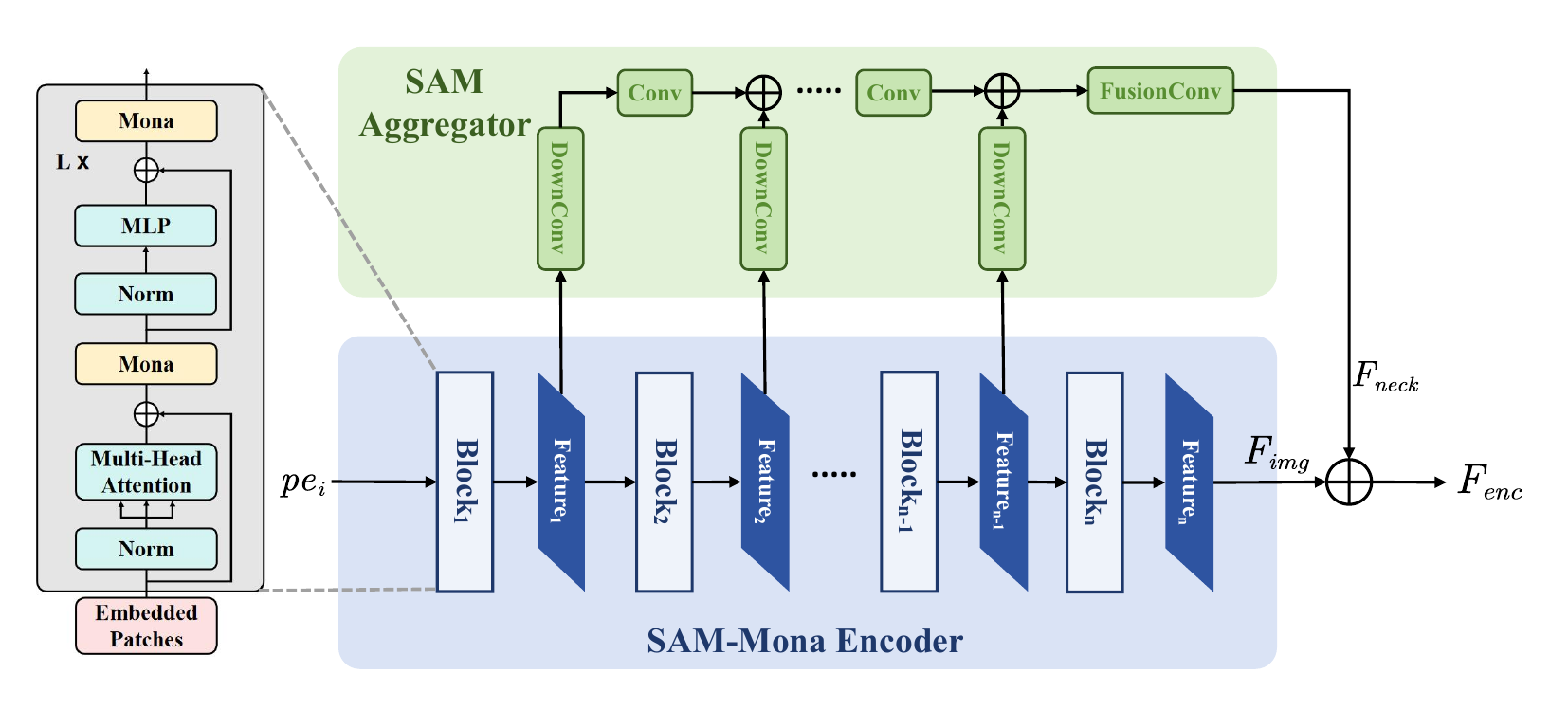}
  \caption{
    The structure of SAM-Mona image encoder.}
  \label{fig_3}
\end{figure*}
The proposed method MC-SAM SEG comprises two main components: an image encoder, which integrates the SAM-Mona encoder and the SAM Aggregate Neck, and a mask decoder that consists of a Pixel Decoder, an additional Transformer Decoder, and two MLP layers, one for mask generation and the other for class prediction.  
The structure of the image encoder is illustrated in Fig. 3, and the computation process can be described as follows:
\begin{equation}
  \begin{aligned}
  F_{\text {img }},\left\{F_i\right\} & =\Phi_{SM Enc}\left(I_i\right) \\[4pt]
  F_{\text {neck }} & =\Phi_{\text {Agg}}\left(\left\{F_i\right\}\right) \\[4pt]
  F_{\text {enc }} & =F_{\text {img }}+F_{\text {neck }}
  \end{aligned}
  \end{equation}

In the SAM-Mona encoder, an input remote sensing image $I{i} \in \mathbb{R}^{h \times w \times c}$ is cropped into N patches $p_i$ and is flattened by a linear layer. These patches are concatenated with position embeddings, resulting in N+1 patch embeddings $pe_i$ which are then fed into the image encoder $\Phi_{SM Enc}$. 
The multi-scale features $\left\{F_i\right\}$ from the intermediate layers of the image encoder are extracted and aggregated 
by the feature aggregator $\Phi_{Agg}$. These aggregated features $F_{neck}$ are added to the encoder's output $F_{img}$, producing the final output of the image encoder $F_{enc}$.
The mask decoder, as illustrated in Fig. 4, can be described by the following computation process:
\begin{equation}
  \begin{aligned}
  F_{\text {dec }},\left\{F_p\right\} & =\Phi_{\text {PixDec }}\left(F_{\text {enc }}\right) \\[4pt]
  F_{\text {multi }} & =\Phi_{\text {TransDec }}\left(\left\{F_p\right\}\right) \\[4pt]
  C_{\text {pre }} & =MLP_{\text {cls }}\left(F_{\text {multi }}\right) \\[4pt]
  M_{\text {pre }} & =MLP_{\text {mask }}\left(F_{\text {multi }} \times F_{\text {dec }}\right)
  \end{aligned}
  \end{equation}

The Multi-scale Deformable Attention Transformer (MSDeformAttn) is employed as the pixel decoder $\Phi_{Pix Dec}$ to transform features $F_{enc}$ into pixel-level results $F_{dec}$. 
The intermediate features  $\left\{F_p\right\}$ of the pixel decoder are fed into a Transformer decoder  $\Phi_{Trans Dec}$ with masked attention to decode them into pixel-level outputs $F_{multi}$. 
$F_{multi}$ includes instance category predictions from the category predictor $MLP_{\text {cls }}$, corresponding to the predicted category of instances $C_{p r e} \in \mathbb{R}^{1 \times q}$.
Additionally, aggregated inputs from $F_{multi}$ and $F_{dec}$ are fed into the mask predictor $MLP_{\text {mask}}$ to predict instance masks $M_{p r e} \in \mathbb{R}^{h \times w \times q}$. 
The parameter $q$ specifies the number of designated tokens used to control the number of masks predicted by the model.
\begin{figure*}[t]
  \centering
  \includegraphics[width=6.5in]{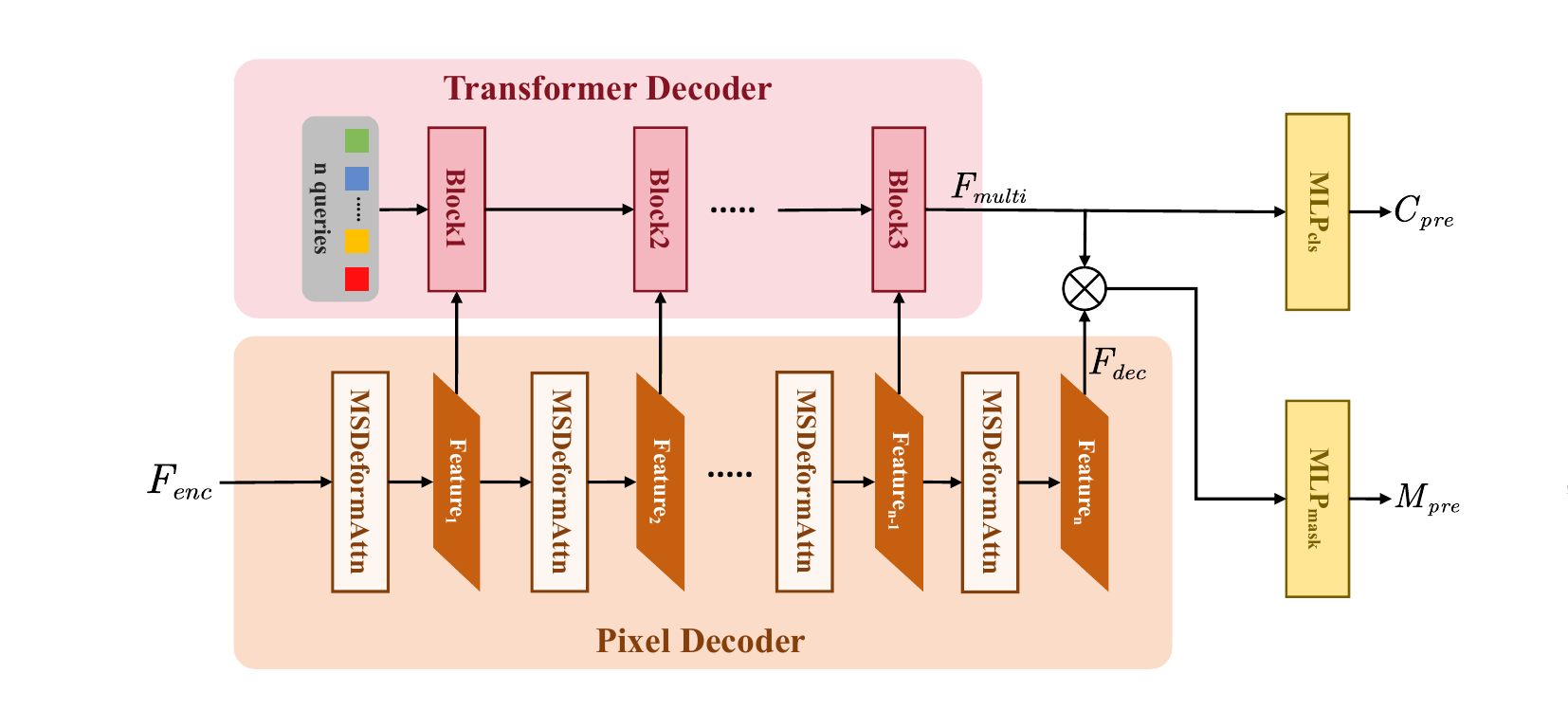}
  \caption{
    The structure of MC-SAM SEG's mask decoder.}
  \label{fig_4}
\end{figure*}

\subsection{Transfer Learning}
As previously mentioned, directly training the full model with the original SAM weights incurs prohibitive computational costs 
and large time consumption. Therefore, partial weights of the SAM-Mona encoder are froze during training, as illustrated in Fig. 5. 
Specifically, only the Mona adapters in the SAM-Mona encoder were allowed to participate in backpropagation. 
This setup enables the Mona adapters to fully engage in the training for remote sensing image instance segmentation, 
providing the encoder with sufficient domain-specific knowledge while preserving robust feature extraction capabilities. 
All other components are fully trained to ensure the generation of high-quality masks. 
This PEFT strategy balances computational efficiency and model performance, 
ensuring that the model adapts effectively to the unique challenges of remote sensing imagery.

The weighted cross-entropy loss and dice loss are conducted to measure the discrepancy between the predicted masks and the ground truth masks. 
The cross-entropy loss evaluates the pixel-wise classification accuracy, ensuring that the predicted masks closely match the true instance masks. 
The dice loss complements this by focusing on the overlap between the predicted and actual masks, 
enhancing the model ability to accurately segment objects, especially in cases of class imbalance.
This ensures that the model not only segments the objects accurately but also assigns the correct category to each instance, 
thereby improving the overall performance and reliability of the instance segmentation process.
The calculation of the loss function can be described as follows:
\begin{equation}
  \begin{aligned}
  \mathcal{L}_{\text {ce-cls}} & =-\frac{1}{N} \sum_i \sum_{c=1}^M y_{ic} \log \left(c_{ic}\right)\\[4pt]
  \mathcal{L}_{\text {ce-seg}} & =-\frac{1}{N} \sum_i \sum_{c=1}^M y_{ic} \log \left(m_{ic}\right)\\[4pt]
  \mathcal{L}_{\text {dice}} & = 1-\frac{2 \sum_1^N y_i m_i+\varepsilon}{\sum_1^N (y_i+m_i)+\varepsilon}\\[4pt]
  \mathcal{L} & =\lambda_{1}\mathcal{L}_{\text {ce-cls}}+\lambda_{2}\mathcal{L}_{\text {ce-seg}}+\lambda_{3}\mathcal{L}_{\text {dice}}
  \end{aligned}
  \end{equation}
where \( y_{ic} \) and \( y_{i} \) is an indicator function that equals 1 if the prediction is correct and 0 otherwise. 
\( c_{ic} \) represents the probability that the predicted category belongs to \( c \), \( M \) is the number of categories.
The predicted masks $M_{p r e} \in \mathbb{R}^{h \times w \times q}$ are sampled to obtain a set of points $m_{ic}\in \mathbb{R}^{n \times q}$ 
for cross-entropy calculation and $m_{i}\in \mathbb{R}^{n \times q}$ for dice coefficient calculation. 
\(\lambda\) is the weight balancing the different loss components.

\section{EXPERIMENTS}
\subsection{Experimental Dataset and Settings}
This paper validates the effectiveness of the proposed method on two public remote sensing instance segmentation datasets: 
the WHU building extraction dataset \cite{ji2018fully} and the high-resolution SAR images dataset (HRSID) \cite{wei2020hrsid}. 
These datasets are widely used for performance evaluation in the field of remote sensing instance segmentation.

\textit{1) WHU}: The experiments utilize the aerial image subset from the WHU Building Extraction Dataset. 
This subset contains 8188 non-overlapping RGB images of 512 × 512 pixels, with a spatial resolution of 0.0075 meters. 
It includes annotations for 220,000 building samples in rural, residential, cultural, and industrial areas of Christchurch, New Zealand. 
Following the settings of RSPrompter, 4736 images are used for the training set, 1036 images for the validation set, 
and 2416 images for the test set. The provided instance segmentation annotations are used for training and testing.

\textit{2) HRSID}: HRSID is used for ship detection, semantic segmentation, and instance segmentation tasks in high-resolution SAR images. 
The experiments utilize the instance segmentation subset of the dataset, 
which consists of 136 panoramic SAR images with resolutions ranging from 1m to 5m. These images are cropped into 800×800 pixel SAR images
with a 25$\%$ overlap, resulting in a total of 5604 cropped SAR images and 16951 ships. The experiments follow the official split, 
using 65$\%$ of the images for training and 35$\%$ for testing.

\subsection{Evaluation Protocol and Metrics}
The experiments utilize the COCO average precision metric to evaluate the performance of our proposed method. 
This metric is widely used in the computer vision field to assess models' performance in tasks such as object detection and instance segmentation. 
An instance's prediction is considered accurate if its bounding box or mask overlaps with the corresponding ground truth by more than a threshold \( T \) and the predicted category matches. 
In this study, the experiments use $\mathrm{AP}_{\text {mask }}$, $\mathrm{AP}_{\text {box }}$, $\mathrm{AP}_{\text {mask }}^{50}$, $\mathrm{AP}_{\text {box }}^{50}$, $\mathrm{AP}_{\text {mask }}^{75}$and  $\mathrm{AP}_{\text {box }}^{75}$ for evaluation.
$\mathrm{AP}$ is the average precision calculated over 10 IoU thresholds (0.50:0.05:0.95) and measures the overall performance of the model across all categories. 
$\mathrm{AP}^{50}$ represents the average precision at an IoU threshold of 0.50, 
which is commonly used and provides an indicator of the model's performance at a standard threshold. 
$\mathrm{AP}^{75}$ corresponds to an IoU threshold of 0.75, 
offering a stricter evaluation as it requires a higher overlap between the predicted and ground truth bounding boxes, 
thus serving as a more challenging task.
\begin{figure*}[t]
  \centering
  \includegraphics[width=6.5in]{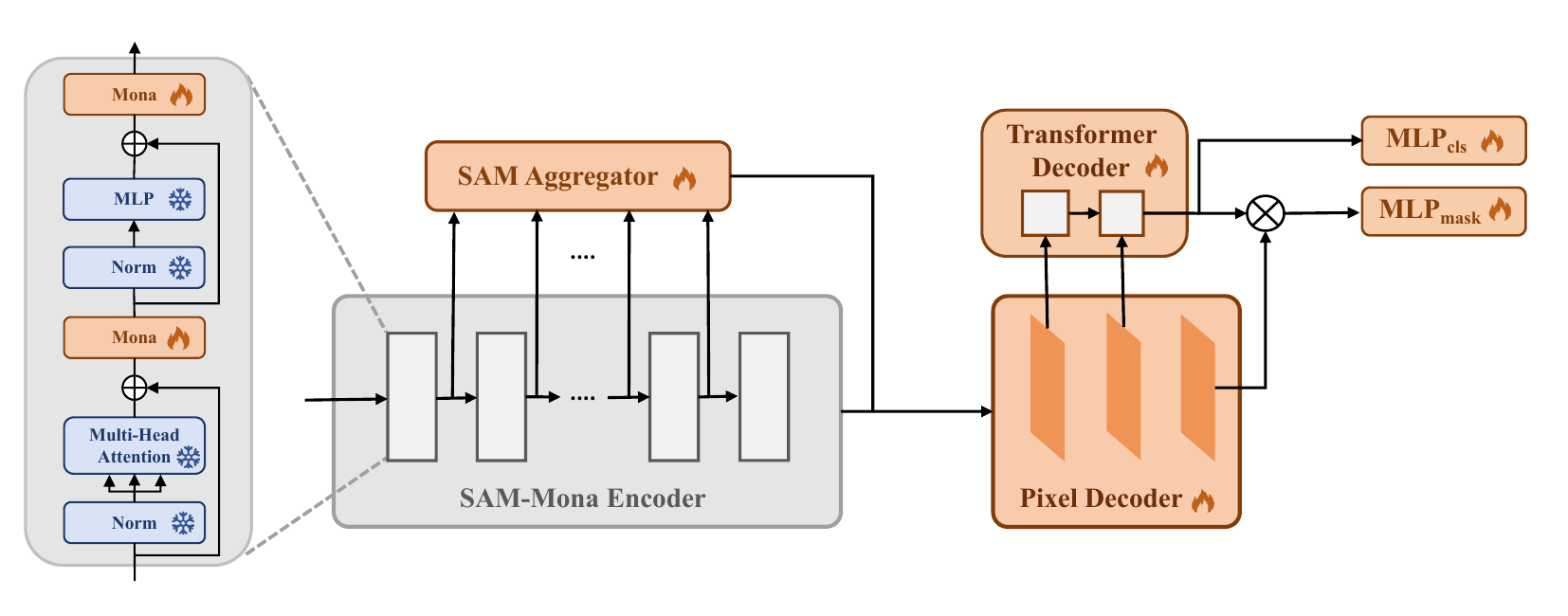}
  \caption{
    Transfer Learning: The blue modules represent frozen components, with non-trainable parameters. 
    The orange modules represent trainable components, where parameters can be fine-tuned. 
    The gray modules indicate partially trainable components.}
  \label{fig_5}
\end{figure*}
\subsection{Implementation Details}
\textit{1) Structural Details}: SAM's official versions provided are base, large, and huge. 
The experiments use the base version of the image encoder in the proposed model, which employs 12 standard Transformer blocks for image encoding. 
A standard Mona adapter is inserted in the proposed model after the multi-head attention and before the final output in each block, 
resulting in a total of 24 Mona adapters being incorporated into the image encoder for transfer learning. 
The feature aggregator follows the settings from RSPrompter \cite{chen2024rsprompter}, maintaining the same hyperparameters. 
For the decoder, the experiments reference the parameter settings from Mask2Former and set different numbers of learnable tokens for each dataset: 
100 tokens for the WHU dataset and 30 tokens for the HRSID dataset.

\textit{2) Training Details}: The model is built on PyTorch and trained and evaluated on two NVIDIA A800 80G GPUs. 
The input to the model are 1024 × 1024 pixel images with a 50$\%$ probability of horizontal flip augmentation. 
During training, the original weights of the SAM image encoder are frozen, 
and only the parameters in the Mona adapters and the decoder are updated via backpropagation. 
The optimizer used is AdamW with an initial learning rate of 1e-4 and a weight decay of 1e-3. 
After a one-epoch warm-up using the LinearLR strategy, the main training employs the CosineAnnealingLR \cite{loshchilov2016sgdr} scheduler. 
The batch size is set to 32×2 (32 per GPU), and training is conducted for 300 epochs on both the WHU and HRSID datasets. 
During testing, the number of predicted instances per image is set to not exceed the number of learnable tokens.

\subsection{Comparison with the State-of-the-Art}
\textit{1) Quantitative Results on WHU Dataset}: The WHU dataset is a single-class building instance segmentation dataset. 
The primary challenges of this dataset include the accurate segmentation of complex large buildings, densely packed buildings, 
and buildings with complex backgrounds.

In the experiments on the WHU dataset, 
the performance of the proposed MC-SAM SEG model and state-of-the-art (SOTA) instance segmentation models used in general visual tasks are compared, 
including Mask R-CNN \cite{cai2018cascade}, Mask Scoring (MS) R-CNN \cite{huang2019mask}, HTC \cite{chen2019hybrid}, SCNet \cite{vu2021scnet}, Mask2Former \cite{cheng2022masked}, CATNet \cite{liu2021catnet}, and SOLO \cite{wang2020solo}. 
Additionally, the experiments compare the proposed model with SOTA instance segmentation models specifically designed for remote sensing images, such as HQ-ISNet \cite{su2020hq} and RSPrompter \cite{chen2024rsprompter}. 
Notably, RSPrompter is also based on the SAM model and achieves excellent performance through prompt engineering.
The experiments conduct a comprehensive evaluation of the proposed MC-SAM SEG model on the optical remote sensing image instance segmentation task using this dataset. 
To assess the impact of the Mona adapter, the experiments also provide the evaluation results of SAM SEG without the Mona adapter. 
Tab. I presents the AP values obtained by different methods on this dataset.

\begin{table*}[!t]
  \caption{The Table Illustrates a Comparison Between the Proposed Method and Other state-of-the-art Methods on the WHU Dataset. 
  it Exhibits AP($\%$) Values for Masks at Distinct IOU Thresholds.\label{tab:table1}}
  \centering
  \setlength{\tabcolsep}{6mm}{
  \begin{tabular}{c|ccc|ccc}
  \toprule
  Method&$\mathrm{AP}_{\text {mask }}$&$\mathrm{AP}_{\text {mask }}^{50}$&$\mathrm{AP}_{\text {mask }}^{75}$&$\mathrm{AP}_{\text {box }}$&$\mathrm{AP}_{\text {box }}^{50}$&$\mathrm{AP}_{\text {box }}^{75}$ \\
  \midrule
  Mask R-CNN            &60.8&84.2&71.4                   &56.1&80.2&	63.3               \\
  MS R-CNN              &53.4&80.1&62.3                   & 54.9&	79.9&	62.8                 \\
  HTC                   &56.7&82.0&68.1                   & 58.6	 &81.5	 &67.3                   \\
  SCNet                 &57.5&81.5&67.2                   &58.4	&81.0	&67.0                                       \\
  Mask2Former           &62.8&85.7&72.7                   & 60.4	&82.7	&68.5                                    \\
  CATNet                &61.1&85.6&70.7                   &59.6	&84.3	&68.5                                      \\
  SOLO                  &60.1&82.0&70.4                   & 51.0	&74.9	&57.2                               \\
  HQ-ISNet              &62.3&86.7&71.5                   & 61.2	&85.0	&68.9                                            \\
  RSPrompter            &69.2&90.2&79.9                   &70.4	&89.2	&79.1                          \\
  \midrule
  SAM SEG               &63.2&87.3&74.9                   &59.7	&85.9	&65.9                                       \\
  MC-SAM SEG&\textbf{71.2}&\textbf{91.5}&\textbf{81.1}                         & \textbf{71.0}&\textbf{91.1}&\textbf{85.6}   \\
  $\empty$&\textbf{+2.0}&\textbf{+1.3}&\textbf{+1.2}&\textbf{+0.6}&\textbf{+1.9}&\textbf{+6.5}\\
  \bottomrule
  \end{tabular}}
  \end{table*}

  \begin{table*}[!t]
    \caption{The Table Illustrates a Comparison Between the Proposed Method and Other state-of-the-art Methods on the HRSID Dataset. 
    it Exhibits AP($\%$) Values for Masks at Distinct IOU Thresholds.\label{tab:table2}}
    \centering
    \setlength{\tabcolsep}{6mm}{
    \begin{tabular}{c|ccc|ccc}
    \toprule
    Method&$\mathrm{AP}_{\text {mask }}$&$\mathrm{AP}_{\text {mask }}^{50}$&$\mathrm{AP}_{\text {mask }}^{75}$&$\mathrm{AP}_{\text {box }}$&$\mathrm{AP}_{\text {box }}^{50}$&$\mathrm{AP}_{\text {box }}^{75}$ \\
    \midrule
    Mask R-CNN&52.6&84.1&61.6  &61.6&86.4&70.2                     \\
    MS R-CNN&54.0&85.6&64.1    &64.9&88.1&73.8                    \\
    HTC&55.2&84.9&66.5         &66.6&86.0&77.1                    \\
    SCNet&54.4&82.4&65.9       &60.7&82.5&72.4                    \\
    Mask2Former&62.1&90.5&79.6 &65.3&85.9&75.0                      \\
    CATNet&62.5&92.1&78.9      &64.7&83.1&75.1\\
    HQ-ISNet&54.6&85.0&65.8    &66.0&86.1&75.6                     \\
    LFG-Net&61.3&87.7&73.6     &65.0&91.2&83.5                     \\
    SISNet&57.9&89.7&70.3      &\textbf{68.5}&90.7&78.9                         \\
    FL-CSE-ROIE&56.9&87.3&67.7 &67.0&89.3&78.5                       \\
    RSPrompter&64.6&94.1&81.7  &63.7&89.9&80.7                       \\
    \midrule
    SAM SEG&59.7&87.9&79.7       &57.5&86.3&67.8                         \\
    MC-SAM SEG&\textbf{66.4}&\textbf{97.0}&\textbf{88.1}   &66.0&\textbf{93.6}&\textbf{84.2}         \\
    $\empty$&\textbf{+1.8}&\textbf{+2.9}&\textbf{+6.4}&-1.5&\textbf{+2.4}&\textbf{+0.7}\\
    \bottomrule
    \end{tabular}}
    \end{table*}

From Tab. I, it can be observed that the SAM SEG, which is trained with frozen SAM weights, 
achieves a $\mathrm{AP}_{\text {mask }}$ performance of 63.2. This performance is only lower than RSPrompter (69.2), 
demonstrating that the original SAM model already possesses a certain level of understanding of remote sensing images, 
even without specialized domain knowledge. The MC-SAM SEG surpasses RSPrompter, achieving the highest performance at 71.2. 
Specifically, MC-SAM SEG shows improvements of 2.0$\%$, 1.3$\%$, 1.2$\%$, 0.6$\%$, 1.9$\%$, and 6.5$\%$, in $\mathrm{AP}_{\text {mask }}$, $\mathrm{AP}_{\text {mask }}^{50}$, $\mathrm{AP}_{\text {mask }}^{75}$, $\mathrm{AP}_{\text {box }}$, $\mathrm{AP}_{\text {box }}^{50}$ and $\mathrm{AP}_{\text {box }}^{75}$metrics, respectively. 
This demonstrates that supplementing SAM with domain-specific knowledge through the Mona adapter effectively enhances the model's understanding in the specialized remote sensing field and improves its performance ceiling.

\textit{2) Quantitative Results on HRSID Dataset}:
HRSID is a single-class SAR ship instance segmentation dataset. SAR images are represented as grayscale images, 
recording the reflection intensity or amplitude of radar signals and are accompanied by speckle noise, 
which makes them significantly different from optical remote sensing datasets.

In the experiments on the HRSID dataset, 
the performance of the proposed model and state-of-the-art (SOTA) instance segmentation models from general vision tasks are compared, such as Mask R-CNN \cite{cai2018cascade}, 
Mask Scoring (MS) R-CNN \cite{huang2019mask}, HTC \cite{chen2019hybrid}, SCNet \cite{vu2021scnet},CATNet \cite{liu2021catnet} and Mask2Former \cite{cheng2022masked}. 
Additionally, the experiments compare MC-SAM SEG model with SOTA instance segmentation models specifically designed for remote sensing images such as HQ-ISNet \cite{su2020hq}, 
LFG-Net \cite{wei2022lfg}, SISNet \cite{shao2023scale}, FL-CSE-ROIE \cite{zhang2022full}, and RSPrompter \cite{chen2024rsprompter}. 
Notably, LFG-Net, SISNet, and FL-CSE-ROIE are specifically designed for SAR image instance segmentation tasks and exhibit superior performance on SAR images. 
The experiments conduct a comprehensive evaluation of proposed MC-SAM SEG model on SAR image instance segmentation tasks using this dataset 
and also provide the evaluation results for SAM SEG without the Mona adapter. 
Tab. II shows the AP values obtained by different methods on this dataset.
\begin{figure*}[t]
  \centering
  % \vspace{-3.5cm} 
  \includegraphics[width=6in]{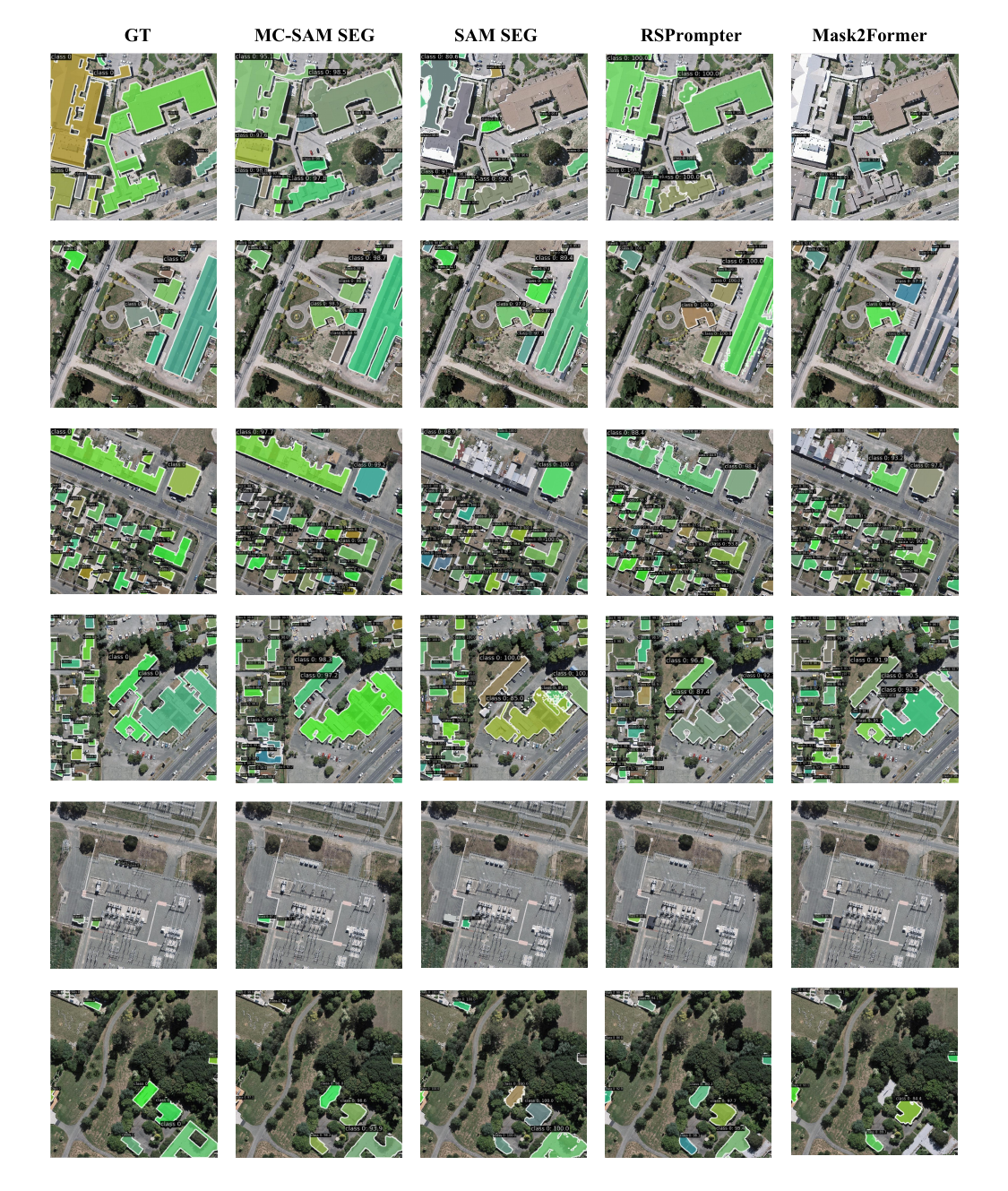}
  \caption{Visualize examples of segmentation results on test images from the WHU dataset, 
  comparing the proposed method with other state-of-the-art methods. 
  Different colors are used to distinguish between different instances.}
  \label{fig_6}
\end{figure*}
\begin{figure*}[t]
  \centering
  % \vspace{-3.5cm} 
  \includegraphics[width=6in]{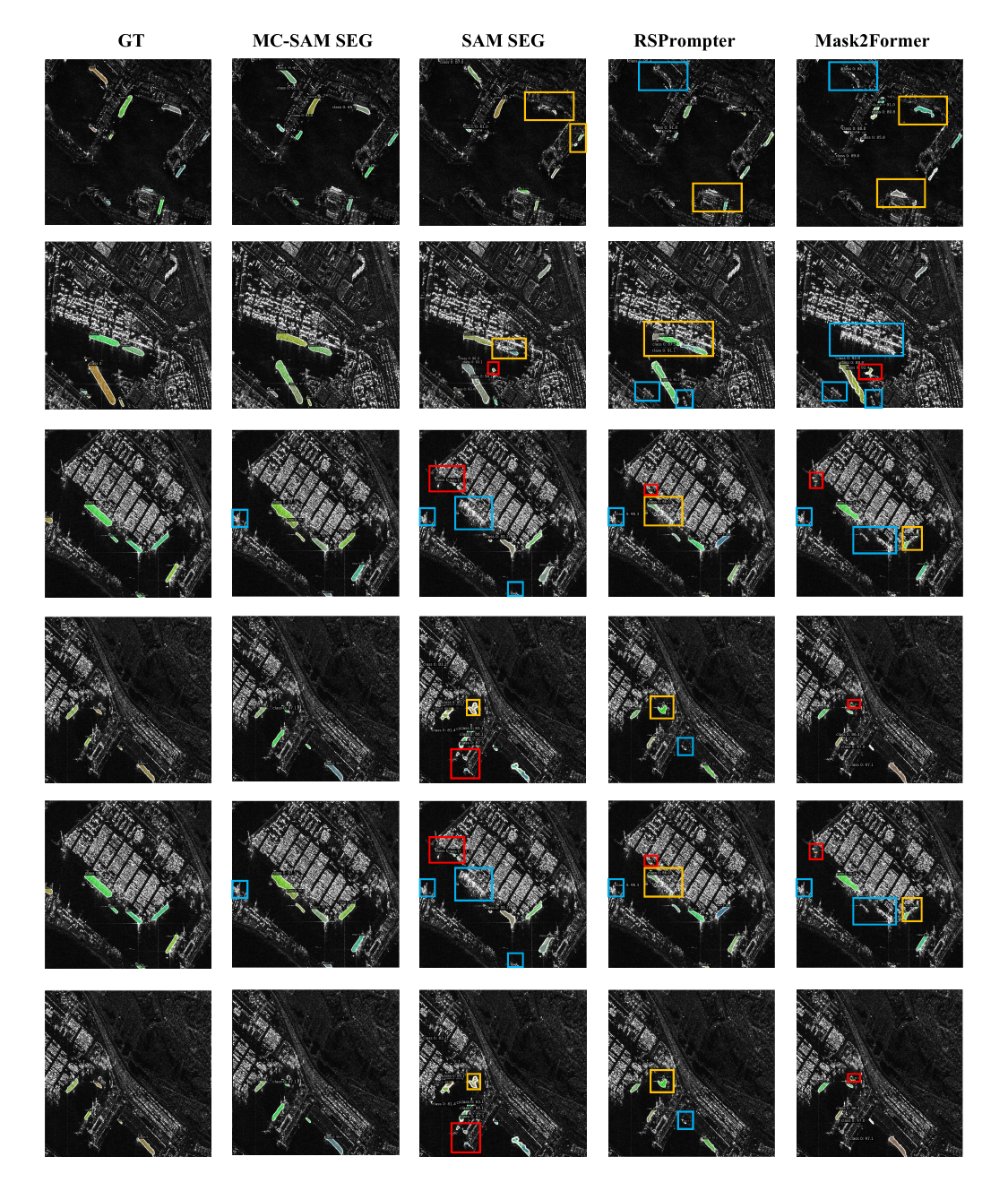}
  \caption{Visualize examples of segmentation results on test images from the HRSID dataset, 
  comparing the proposed method with other state-of-the-art methods. 
  Different colors represent different instances, with red boxes marking false positives, blue boxes marking missed targets, 
  and yellow boxes marking poorly segmented targets.}
  \label{fig_7}
\end{figure*}

From Tab. II, it can be observed that due to the differences between SAR and optical images, 
general vision task instance segmentation models such as Mask R-CNN (52.2), Mask Scoring (MS) R-CNN (54.9), HTC (55.2), and SCNet (54.4) 
generally perform worse than models designed with SAR domain-specific knowledge like LFG-Net (61.3), SISNet (57.8) and FL-CSE-ROIE (56.9). 
Notably, models such as LFG-Net (60.2), Mask2Former (61.1), and RSPrompter (65.4) outperform SAM SEG (59.7) on this dataset. 
However, it is noteworthy that the SAM SEG, with the SAM image encoder weights frozen during training, 
performs slightly lower than SOTA methods, demonstrating that SAM's original weights already possess some understanding of SAR images. 
After incorporating SAR domain knowledge through training with the SAM-Mona encoder, 
the MC-SAM SEG model shows improvements of 1.8$\%$, 2.9$\%$, 6.4$\%$, 2.4$\%$, and 0.7$\%$ in $\mathrm{AP}_{\text {mask }}$, $\mathrm{AP}_{\text {mask }}^{50}$, $\mathrm{AP}_{\text {mask }}^{75}$, $\mathrm{AP}_{\text {box }}^{50}$ and $\mathrm{AP}_{\text {box }}^{75}$ metrics respectively over other SOTA methods. 
This indicates that using Mona Tuning to supplement SAM with SAR-specific knowledge can effectively enhance the model's understanding 
and performance in remote sensing field, thus raising the performance ceiling.

\textit{3) Qualitative Visual Comparisons}:
To visually observe the performance differences between models and verify the effectiveness of the proposed method, 
The segmentation results of the proposed method are visualized alongside other advanced methods for qualitative analysis. 
Fig. 6 and Fig. 7 show the visualization results on the WHU and HRSID datasets, respectively. 
To provide a clearer presentation, segmentation results with a prediction confidence greater than 0.75 are selected for display.

As shown in Fig. 6, comparing the proposed model MC-SAM SEG with SAM SEG, RSPrompter, and Mask2Former, 
it can be observed that MC-SAM SEG produces more accurate, complete, 
and clearer instance segmentation results in scenarios involving large complex buildings, densely packed houses, 
and buildings with complex backgrounds, compared to other state-of-the-art methods.

As shown in Fig. 7, comparing the proposed model MC-SAM SEG with SAM SEG, RSPrompter, and Mask2Former, 
Red boxes are used to highlight false positives, blue boxes for missed detections, and yellow boxes for suboptimal segmentation results for clarity. 
It can be observed that, compared to other state-of-the-art methods, 
MC-SAM SEG more accurately distinguishes between background and ship instances in SAR ship instance segmentation tasks, 
reducing false positives and missed detections while providing higher quality segmentation results.

\subsection{Ablation Study}
To verify the effectiveness of each component of the proposed method, 
ablation experiments are conducted on the HRSID dataset. Specifically, 
the experiments analyze the impact of different configurations of the image encoder on the model's performance.

\begin{table}[!t]
  \caption{The Table Illustrates the Impact of SAM and Mona Tuning in the Instance Segmentation Performance on the HRSID Dataset. 
  it Exhibits AP($\%$) Values for Masks at Distinct IOU Thresholds.\label{tab:table3}}
  \centering
  \setlength{\tabcolsep}{3mm}{
  \begin{tabular}{cc|ccc}
  \toprule
  SAM&Mona Tuning&$\mathrm{AP}_{\text {mask }}$&$\mathrm{AP}_{\text {mask }}^{50}$&$\mathrm{AP}_{\text {mask }}^{75}$ \\
  \midrule
  $\empty$&$\empty$&45.4&74.3&56.1  \\
  $\empty$&\checkmark&56.3&84.2&65.0  \\
  \checkmark&$\empty$&59.7&87.9&79.7  \\
  \checkmark&\checkmark&\textbf{66.4}&\textbf{97.0}&\textbf{88.1}  \\
  \bottomrule
  \end{tabular}}
  \end{table}

  \textit{1) Impact of Image Encoder}:  The SAM image encoder is a Vision Transformer (ViT) pretrained using Mask Auto Encoder (MAE), 
    known for its powerful zero-shot inference capabilities due to training on extensive datasets. 
    However, the lack of remote sensing data in the pretraining phase limits SAM's performance in this specific field. 
    To address this, experiments comparing different configurations of the image encoder are set up to evaluate the impact of SAM pretraining weights and Mona Tuning on model performance.
    The following scenarios are compared:
\begin{itemize}
  \item{ViT with traditional pre-trained weights, loaded and frozen.}
  \item{ViT with traditional pre-trained weights, loaded and fine-tuned using Mona.}
  \item{ViT with SAM weights, loaded and frozen.}
  \item{ViT with SAM weights, loaded and fine-tuned using Mona.}
\end{itemize}

The performance of each configuration is measured and compared, as shown in Tab. III.
The metrics include $\mathrm{AP}_{\text {mask }}$, $\mathrm{AP}_{\text {mask }}^{50}$, $\mathrm{AP}_{\text {mask }}^{75}$.
The performance of traditional pre-trained weights is the lowest among the configurations, 
indicating that traditional weights lack the specialized knowledge required for remote sensing image segmentation.
Without fine-tuning, using SAM model weights provided a significant 14.3$\%$ performance improvement over traditional weights, 
highlighting SAM's superior general image understanding capabilities even without domain-specific training,
indicating that SAM has a stronger image understanding capability compared to traditional pre-trained weights. 
After fine-tuning with Mona, SAM weights achieves the highest performance across all metrics, 
further validating the effectiveness of Mona Tuning in leveraging SAM's robust foundational capabilities and enriching them with remote sensing domain knowledge.
The ablation experiments confirm the following:
\begin{itemize}
  \item{SAM's pre-trained weights provide a strong foundation for image understanding, 
significantly outperforming traditional pre-trained weights in remote sensing tasks.}
\item{Mona tuning enhances both traditional and SAM pre-trained weights by incorporating domain-specific knowledge, 
with the combination of SAM weights and Mona Tuning yielding the best results.}
\end{itemize}

\textit{2) Impact of Different PEFT Techniques}:
In this section, the experiments evaluate the impact of different Parameter-Efficient Fine-Tuning (PEFT) techniques on the performance of the proposed model. 
The experiments compare the effectiveness of Mona, Adapter, and LoRA in enhancing the model's capabilities for remote sensing image instance segmentation.
Adapter and Mona are insertred into the image encoder in the same manner,
and LoRA is employed to the linear and convolutional layers of the image encoder, following the official setup.
The results are presented in Tab. IV, showing AP($\%$) values for masks at different IoU thresholds.
The comparison of different PEFT techniques indicates that
Mona is highly effective for remote sensing image instance segmentation, significantly enhancing the model's performance.
Adapter also improves the model's performance but to a lesser extent than Mona.
LoRA does not perform as well as the other techniques, possibly due to its initial design for language models rather than visual tasks.
These results demonstrate that Mona's design, specifically tailored for visual tasks, 
makes it more suitable for the task of remote sensing image instance segmentation, 
effectively enhancing the SAM model's capabilities in this domain.
\begin{table}[!t]
  \caption{The Table Illustrates a Comparison Between Different PEFT Methods on the HRSID Dataset. 
  it Exhibits AP($\%$) Values for Masks at Distinct IOU Thresholds.\label{tab:table4}}
  \centering
  \setlength{\tabcolsep}{3mm}{
  \begin{tabular}{ccccc}
  \toprule
  Model&PEFT&$\mathrm{AP}_{\text {mask }}$&$\mathrm{AP}_{\text {mask }}^{50}$&$\mathrm{AP}_{\text {mask }}^{75}$ \\
  \midrule
  \multirow{4}*{SAM SEG}   &/                  &59.7          &87.9            &79.7  \\
  ~                        &Adapter            &63.1          &93.6            &86.0  \\
  ~                        &LoRA               &58.7          &90.1            &71.4  \\
  ~                        &Mona               &\textbf{66.4} &\textbf{97.0}   &\textbf{88.1}  \\
  \bottomrule
  \end{tabular}}
  \end{table}
\section{CONCLUSION}
In this paper, MC-SAM SEG is proposed to adapt Segment Anything Model to effective remote sensing image instance segmentation without relying on input prompts. 
The integration of Multi-Cognitive Visual Adapters in SAM image encoder named SAM-Mona encoder is implemented to achieve feature construction of remote sensing images. 
The output image features of SAM-Mona encoder are then aggregated through an aggregator and processed by a mask decoder, 
which consists of a pixel decoder and transformer decoder to generate high-quality instance segmentation results.
Ablation experiments validate the contributions of each component within MC-SAM SEG, 
particularly highlighting the impact of Mona adapters and the effectiveness of PEFT technique.
MC-SAM SEG achieves outstanding performance on both the WHU and HRSID datasets, and enhances the advanced capabilities of Segment Anything Model in remote sensing image instance segmentation task.

\bibliographystyle{ieeetr}
\balance
  \bibliography{reference}
  
\end{document}